\pdfoutput=1
\documentclass[11pt]{article}

\usepackage{acl}

\usepackage{times}
\usepackage{latexsym}

\usepackage[T1]{fontenc}
\usepackage[utf8]{inputenc}

\usepackage{microtype}

\usepackage{placeins}
\usepackage{xcolor}
\usepackage{comment}
\usepackage{amsmath}
\usepackage{graphicx}

\newcommand{\best}[1]{\textbf{#1}}
\newcommand{\sbest}[1]{\underline{#1}}

\newcommand{\retwog}{$\text{Re}^2\text{G}$}
\newcommand{\kgi}[1]{$\text{KGI}_{#1}$}

\title{Re$^2$G: Retrieve, Rerank, Generate}
\author{
    Michael Glass\textsuperscript{\rm 1},
    Gaetano Rossiello\textsuperscript{\rm 1},
    Md Faisal Mahbub Chowdhury\textsuperscript{\rm 1},\
    \\
    \textbf{Ankita Rajaram Naik}\textsuperscript{\rm 1 \rm2},
    \textbf{Pengshan Cai}\textsuperscript{\rm 1 \rm2},
    \textbf{Alfio Gliozzo}\textsuperscript{\rm 1}
    \\
    \textsuperscript{\rm 1} IBM Research AI, Yorktown Heights, NY, USA\\
    \textsuperscript{\rm 2} University of Massachusetts Amherst, MA, USA
}

\begin{document}

\maketitle

%August 30, 2021: Abstracts due at 11:59 PM UTC-12
%September 8, 2021: Full papers due at 11:59 PM UTC-12
%September 11, 2021: Supplementary material and code due by 11:59 PM UTC-12

%Submissions may consist of up to 7 pages of technical content plus up to 2 additional pages solely for references.
%All authors must complete a reproducibility checklist.
%All authors are expected to be available to review (light load), unless extenuating circumstances apply.

\begin{abstract}
As demonstrated by GPT-3 and T5, transformers grow in capability as parameter spaces become larger and larger. %GPT-3 and T5 are able to reach effective performance on many tasks.
However, for tasks that require a large amount of knowledge, non-parametric memory allows models to grow dramatically with a sub-linear increase in computational cost and GPU memory requirements.
Recent models such as RAG and REALM have introduced retrieval into conditional generation.
These models incorporate neural initial retrieval from a corpus of passages. %Neural reranking has already proven successful for other tasks requiring retrieval.
%We combine both neural initial retrieval and reranking into a BART-based sequence-to-sequence generation.
We build on this line of research, proposing \retwog{}, which combines both neural initial retrieval and reranking into a BART-based sequence-to-sequence generation.
Our reranking approach also permits merging retrieval results from sources with incomparable scores, enabling an ensemble of BM25 and neural initial retrieval.
To train our system end-to-end, we introduce a novel variation of knowledge distillation to train the initial retrieval, reranker and generation using only ground truth on the target sequence output.
We find large gains in four diverse tasks: zero-shot slot filling, question answering, fact checking and dialog, with relative gains of 9\% to 34\% over the previous state-of-the-art on the KILT leaderboard.
% (77.05-70.58)/70.58 = 0.09166902805327287
% (49.80-37.91)/37.91 = 0.3136375626483778
% (61.78-46.19)/46.19 = 0.3375189434942629
% (78.53-64.41)/64.41 = 0.2192206179164727
% (12.98-11.79)/11.79 = 0.1009329940627652
% Relative to current (4/29/2022) leaderboard
% (77.05-70.58)/70.58 = 0.09166902805327287
% (49.80-44.40)/44.40 = 
% (61.78-54.99)/54.99 =
% (78.53-71.28)/71.28 = 
% (12.98-13.39)/13.39 = 
We make our code available as open source\footnote{\url{https://github.com/IBM/kgi-slot-filling/tree/re2g}}.
\end{abstract}

\section{Introduction}

GPT-3~\cite{gpt3} and T5~\cite{t5} are arguably the most powerful members in a family of deep learning NLP models called transformers. Such models store surprising amount of world knowledge. They have been shown to produce good performance on a range of demanding tasks, especially in generating human like texts. However, such large transformers' capability is tied to the increasingly larger parameter spaces on which they are trained.

Recently, there has been work towards transformers that make use of non-parametric knowledge. REALM (Retrieval Augmented Language Model)~\cite{realm} and RAG (Retrieval Augmented Generation)~\cite{rag} both use an indexed corpus of passages to support conditional generation. By using the corpus as a source of knowledge these models can extend the information available to the model by tens or even hundreds of gigabytes with a sub-linear scaling in computation cost.

These recent advancements, in turn, have been inspired by BART (Bidirectional and Auto-Regressive Transformer)~\cite{bart} that combines a Bidirectional Encoder (e.g. BERT~\cite{bert}) with an Autoregressive decoder (e.g. GPT~\cite{gpt3}) into one sequence-to-sequence model.

We build on this line of research, pioneered by REALM and RAG, and propose a new approach that we call \retwog{} (\textbf{Re}trieve, \textbf{Re}rank, \textbf{G}enerate), which combines both neural initial retrieval and reranking into a BART-based sequence-to-sequence generation. 

There are two particular aspects on which our approach is different from the previous works.
Firstly, our reranking approach permits merging retrieval results from sources with incomparable scores, e.g. enabling an ensemble of BM25 and neural initial retrieval. Secondly, to train our system end-to-end, we introduce a novel variation of knowledge distillation to train the initial retrieval, reranker and generation using only ground truth on the target sequence output. 

The KILT benchmark~\cite{kilt} has been recently introduced to evaluate the capabilities of pre-trained language models to address NLP tasks that require access to external knowledge.
We evaluate on four diverse tasks from KILT: slot filling, question answering, fact checking and dialog. Figure~\ref{fig.taskexamples} shows examples of these tasks. \retwog{} makes significant gains on all four tasks, reaching the top of the KILT leaderboards and establishing a new state-of-the-art. 

The contributions of this work are as follows:
\begin{itemize}
    \item We introduce \retwog{}, demonstrating the effectiveness of reranking for generative language models that incorporate retrieval.
    \item We further extend \retwog{} by ensembling initial retrieval methods, combining neural and traditional keyword-based approaches.
    %\item \retwog{} improves the current state-of-the-art on T-REx for slot filling (+6.47\% KILT-F1), FEVER for fact checking (+14.12\% KILT-AC), and Wizard of Wikipedia for dialog (+1.19\% KILT-F1).
    \item \retwog{} improves the current state-of-the-art of 9\%, 31\%, 34\%, 22\% and 10\% relative gains on the headline KILT metrics for T-REx (slot filling), Natural Questions (question answering), TriviaQA (question answering), FEVER (fact checking), and Wizard of Wikipedia (dialog), respectively. 
    \item We publicly release our code as open source to support continued development.
\end{itemize}

\begin{figure*}
\begin{small}
\noindent\begin{minipage}[t]{0.28\linewidth}%
  \centering
  \fcolorbox{blue}{blue!10}{\parbox{\linewidth\fboxsep\fboxrule}{T-REx}}
  \fbox{\begin{minipage}[t]{\linewidth\fboxsep\fboxrule}%
    \textbf{Input:}\\ Dracula {\small [SEP]} narrative location\\
    \textbf{Output:}  Transylvania\\
    \textbf{Provenance:}  7923-2
  \end{minipage}}\vspace{0.2cm}
  \fcolorbox{blue}{blue!10}{\parbox{\linewidth\fboxsep\fboxrule}{Natural Questions}}
  \fbox{\begin{minipage}[t]{\linewidth\fboxsep\fboxrule}%
    \textbf{Input:} when did bram stoker's dracula come out\\
    \textbf{Output:}  1897\\
    \textbf{Provenance:}  7923-1
  \end{minipage}}\vspace{0.2cm}
  \fcolorbox{blue}{blue!10}{\parbox{\linewidth\fboxsep\fboxrule}{FEVER}}
  \fbox{\begin{minipage}[t]{\linewidth\fboxsep\fboxrule}%
    \textbf{Input:} Dracula is a novel by a Scottish author.\\
    \textbf{Output:} REFUTES\\
    \textbf{Provenance:} 7923-1
  \end{minipage}}
\end{minipage}\hfill
\begin{minipage}[t]{0.38\linewidth}%
  \centering
  \fbox{\begin{minipage}[t]{\linewidth\fboxsep\fboxrule}%
    \textbf{Dracula {\small (7923)}}\\
    
    Dracula is an 1897 Gothic horror novel by Irish author Bram Stoker. It introduced the character of Count Dracula, and established many conventions of subsequent vampire fantasy.\\
      
    The novel tells the story of Dracula's attempt to move from Transylvania to England so that he may find new blood and spread the undead curse, and of the battle between Dracula and a small group of men and a woman led by Professor Abraham Van Helsing.
    %``Dracula'' has been assigned to many literary genres including vampire literature, horror fiction, the gothic novel, and invasion literature. The novel has spawned numerous theatrical, film, and television interpretations.
  \end{minipage}}
 \end{minipage}\hfill
 \begin{minipage}[t]{0.3\linewidth}%
  \centering
  \fcolorbox{blue}{blue!10}{\parbox{\linewidth\fboxsep\fboxrule}{Wizard of Wikipedia}}
  \fbox{\begin{minipage}[t]{\linewidth\fboxsep\fboxrule}%
    \textbf{Input:} 
    \begin{itemize}
      \item I really like vampires!!
      \item Vampires are intense and based on European folklore. Do you have any favorite vampires?
      \item I think dracula is the best one!!! 
    \end{itemize}
    \textbf{Output:} He's one of the best! He's based on the character from the 1897 horror book of the same name.\\
    \textbf{Provenance:} 7923-1
  \end{minipage}}

 \end{minipage}
 \caption{KILT tasks of slot filling, question answering, fact checking and dialog}
 \label{fig.taskexamples}
\end{small}
\end{figure*}

\section{Related Work}\label{sec.relatedwork}
% Introduction of KILT
%The main advantage of KILT relies on a unified evaluation framework that realigns all the original tasks in the suite using the same dump of Wikipedia, as well as introducing a set of metrics to assess the quality of both the retrieval and end-to-end models.
%\gaetano{Move this first paragraph in the Introduction}
The KILT benchmark and public leaderboard\footnote{\url{https://eval.ai/web/challenges/challenge-page/689/leaderboard}} combines eleven datasets across five tasks. The main advantage of the KILT distribution of these datasets is that the provenance information from each dataset is realigned to reference the same snapshot of Wikipedia.  A unified evaluation script and set of metrics is also provided. In this work, we focus on four tasks, such as Slot Filling~\cite{zsre, trex}, Question Answering~\cite{naturalquestions, triviaqa}, Fact Checking~\cite{DBLP:conf/naacl/ThorneVCM18,DBLP:journals/corr/abs-1811-10971}, and Dialog~\cite{DBLP:conf/iclr/DinanRSFAW19} (see Figure~\ref{fig.taskexamples}).

% KILT vanilla baselines
A set of baseline methods have been proposed for KILT.
GENRE~\cite{genre} is trained on BLINK~\cite{blink} and all KILT tasks jointly using a sequence-to-sequence language model to generate the title of the Wikipedia page where the answer can be found. This method is a strong baseline to evaluate the retrieval performance, but it does not address the downstream tasks. On the other hand, generative models, such as BART~\cite{bart} and T5~\cite{t5}, show interesting performance when fine-tuned on the downstream tasks relying only on the implicit knowledge stored in the weights of the neural networks, without the use of any explicit retrieval component. 

\begin{figure*}[thb]
   \centering
   \includegraphics[width=0.9\linewidth]{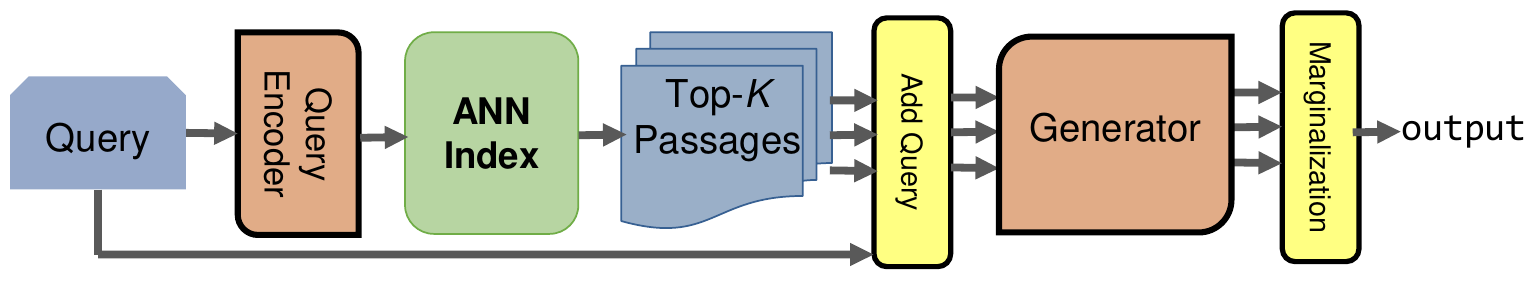}
   \caption{RAG Architecture}
   \label{fig.kgi}
\end{figure*}
\begin{figure*}[bth!]
   \centering
   \includegraphics[width=0.9\linewidth]{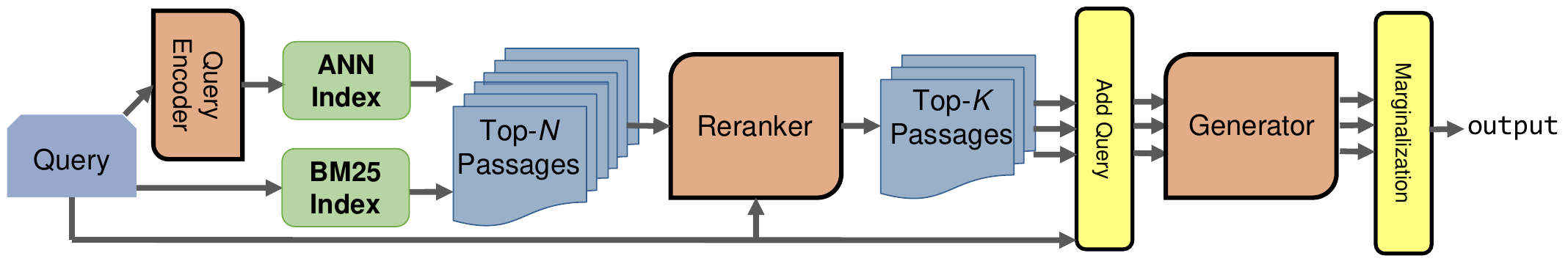}
   \caption{\retwog{} Architecture}
   \label{fig.re2g}
\end{figure*}

% KILT retrieval-based models
RAG~\cite{rag}, an end-to-end retrieval-based generative model, is the best performing baseline in KILT and it incorporates DPR~\cite{dpr} to first retrieve relevant passages for the query, then it uses a model initialized from BART~\cite{bart} to perform a sequence-to-sequence generation from each evidence passage concatenated with the query in order to generate the answer.  Figure~\ref{fig.kgi} shows the architecture of RAG.

Multi-task DPR~\cite{multidpr} exploits multi-task learning by training both DPR passage and query encoder on all KILT tasks.
DensePhrases~\cite{densephrases} addresses the knowledge intensive tasks with a short answer, such as slot filling. It indexes the phrases in the corpus that can be potential answers. The extracted phrases are represented by their start and end token vectors from the final layer of a transformer initialized from SpanBERT~\cite{spanbert}.

Knowledge Graph Induction (KGI)~\cite{kgi_emnlp} combines DPR and RAG models, both trained with task and dataset specific training. %The models are initialized from the Natural Questions~\cite{naturalquestions} trained models for DPR and RAG available from HuggingFace~\cite{huggingface}.
KGI employs a two phase training procedure: first training the DPR model, i.e. both the query and context encoder, using the KILT provenance ground truth. Then, KGI trains the sequence-to-sequence generation and further trains the query encoder using only the target output as the objective. 
%The same query encoder component is trained in both phases. 
This results in large improvements in retrieval performance and, as a consequence, in the downstream tasks. 
%This suggests that designing an even more effective retrieval component could be beneficial to further improve the quality of the generation.

KILT-WEB 2~\cite{kilt_web2} addresses the KILT tasks by broadening the knowledge source used. Rather than rely only on KILT's Wikipedia snapshot, KILT-WEB 2 creates {\sc Sphere} as a knowledge source. {\sc Sphere} is built from CCNet~\cite{ccnet} and over twenty times the size of the Wikipedia corpus. It can use either BM25 or DPR retrieval (though not both combined) followed by a `reader' component, but not trained end-to-end.  The reader component is the Fusion-in-Decoder \cite{fusionInDecoder} model, where retrieved documents are encoded independently, then their encoded representations are concatenated for the decoder.

% check these citations: "We train DPRWEB by finetuning a PAQ-based (Lewis et al., 2021b) bi-encoder checkpoint (Oguz ˘et al., 2021) for 40 epochs on 16 GPUs"

SEAL~\cite{seal_kilt} introduces a novel generative approach to retrieval. Rather than generating the unique document identifier like GENRE, SEAL can generate any ngrams present in the corpus, which are then mapped to passages.  The neural retrieval generator is based on BART and constrained to generate ngrams that appear in the corpus with an FM-Index \cite{FMIndex}. Like KILT-WEB 2, SEAL uses Fusion-in-Decoder as the component responsible for generating the output conditioned on the retrieved passages.

% Related work on reranking
Multi-stage or cascade approaches to retrieval have received ample attention in Information Retrieval (IR) research. The multi-stage approach begins with the initial retrieval phase, where an initial set of documents or passages form the pool of candidates to be considered for ranking. Then one or more phases of increasingly computationally demanding rerankers are applied. Early approaches in learning to rank \cite{learningToRank} used features and linear classifiers. Pre-trained language models, especially BERT \cite{bert}, have shown state-of-the-art performance when applied to the task of relevance ranking. Transformers may be applied as classifiers to each query and passage pair independently~\cite{rerankBERT} or as generators to produce labels for passages in a sequence-to-sequence model~\cite{rerankT5}.

% The best performing of these is . The model  In the baseline RAG approach only the query encoder and generation component are fine-tuned on the task. The passage encoder, trained on Natural Questions \cite{naturalquestions} is held fixed.
% Interestingly, while it gives the best performance of the baselines tested on the task of producing slot fillers, its performance on the retrieval metrics is worse than BM25 \cite{kilt}. This suggests that fine-tuning the entire retrieval component could be beneficial. 
% %the slot filling task is close to factoid QA - so the pre-trained rag model is a good starting point
% Another baseline in KILT is BART$_{LARGE}$ fine-tuned on the slot filling tasks but without the usage of the retrieval model.
% %, also introduced by \citet{kilt}, is an approach to slot filling using the implicit knowledge in the parameters of a pre-trained language model. The pre-trained BART model, is fine-tuned to generate tail objects, conditioned on the head entity and slot, but without any retrieved evidence passage.

%The top-3 passages returned by the resulting passage index were then combined into a single sequence with the query and a BART model was used to produce the answer. This resulted in large gains in retrieval performance.

\section{Methodology}
The approach of RAG, Multi-DPR, and \kgi{} is to train a neural IR (Information Retrieval) component and further train it end-to-end through its impact in generating the correct output.  Figure \ref{fig.kgi} illustrates the end-to-end RAG system.

It has been previously established that results from initial retrieval can be greatly improved through the use of a reranker \cite{learningToRank, wang2011cascade}. Therefore we hypothesized that natural language generation systems incorporating retrieval can benefit from reranking.

In addition to improving the ranking of passages returned from DPR, a reranker can be used after merging the results of multiple retrieval methods with incomparable scores.  For example, the scores returned by BM25 \cite{bm25} are not comparable to the inner products from DPR. Using the scores from a reranker, we can find the top-k documents from the union of DPR and BM25 results.  Figure \ref{fig.re2g} illustrates our extension of RAG with a reranker. We call our system \retwog{} (\textbf{Re}trieve, \textbf{Re}rank, \textbf{G}enerate).

\subsection{Reranker}
\begin{figure}
  \centering
   \includegraphics[width=0.99\linewidth]{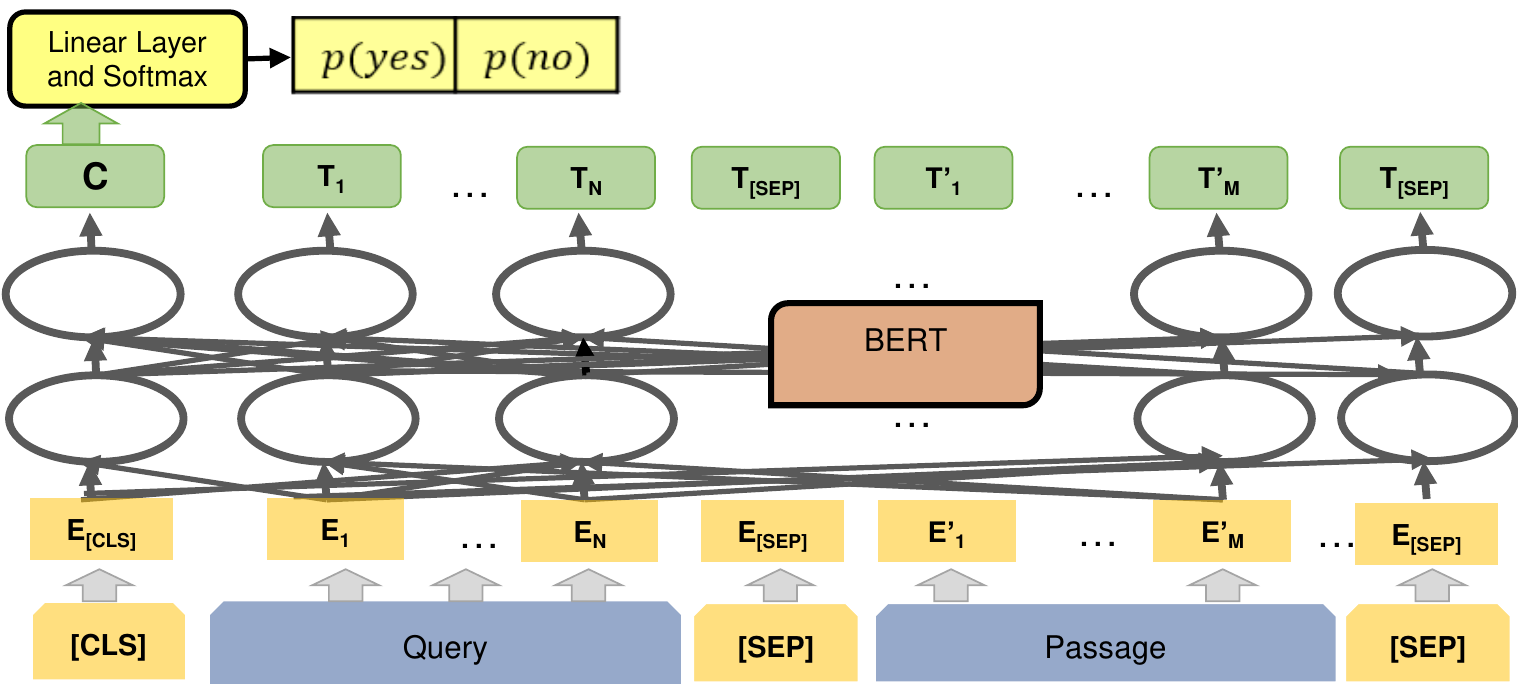}
   \caption{Interaction Model Reranker}
   \label{fig.interaction}
\end{figure}
\begin{figure}
  \centering
   \includegraphics[width=0.99\linewidth]{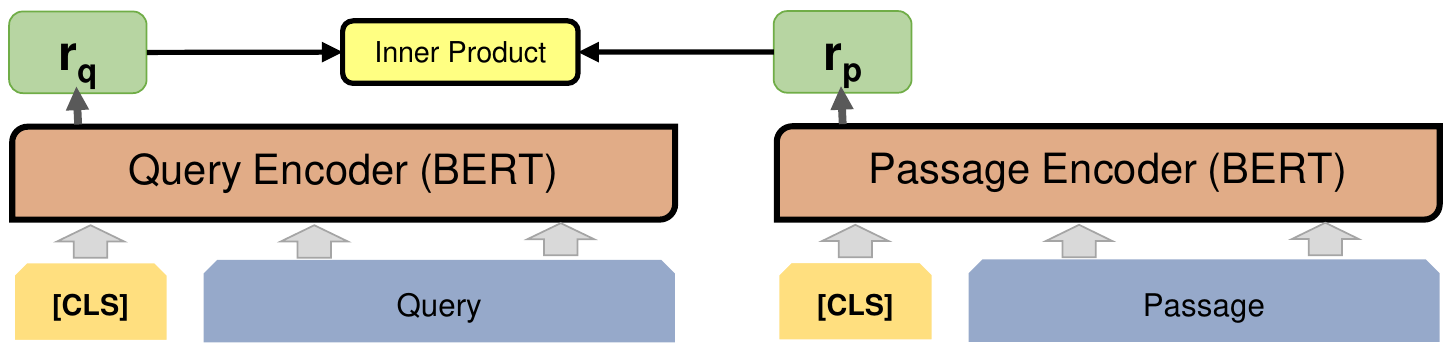}
   \caption{Representation Model for Initial Retrieval}
   \label{fig.representation}
\end{figure}

The reranker we use is based on the sequence-pair classification of \citet{rerankBERT}. This model is shown in Figure \ref{fig.interaction}. The query and passage are input together to a BERT \cite{bert} transformer. Cross attention is applied over the tokens of both sequences jointly. This is called an interaction model.

This model contrasts with the representation model used for initial retrieval. Figure \ref{fig.representation} shows the bi-encoder representation model for DPR. The representation vectors for the query and passage are produced independently. This allows for efficient retrieval by pre-computing vectors for all passages in the corpus and indexing them with an ANN (Approximate Nearest Neighbors) index.
By using an interaction model to rerank the top-N passages from the representation model, we can get the advantages of both model types: accuracy and scalability.

We initialize the reranker from the BERT model trained on MS MARCO \cite{msmarco} by NBoost \cite{nboostRerank} and available through Hugging Face\footnote{\url{https://huggingface.co/nboost/pt-bert-base-uncased-msmarco}}.

%Train KGI0, train reranker over initial retrieval on train set, train Re2G
\subsection{Training}

As Figure \ref{fig.taskexamples} illustrates, KILT tasks are provided with two types of ground truth: the target output sequence and the provenance information indicating the passage or passages in the corpus that support the output.

Our training is carried out in four phases: DPR training, generation training, reranking training, and full end-to-end training.  The initial DPR and reranking phases make use of the provenance ground truth. The generation and full end-to-end training make use of only the target output. 

% CONSIDER: explain dependencies between the phases, maybe a diagram of the training process
Formally:
\begin{itemize}
    \item The original KILT instances are a tuple: $\langle q, t, \mathbf{Prov} \rangle$ where $q$ is the input or prompt, $t$ is the target output, and $\mathbf{Prov}$ is the set of provenance passages that support the target output.
    \item DPR training is a tuple: $\langle q, p^+, p^- \rangle$ where $p^+ \in \mathbf{Prov}$ and $p^-$ where $p^- \in \text{BM25}(q) \wedge p^- \notin \mathbf{Prov}$
    \item Reranking training begins with the application of DPR and BM25, producing tuples:  $\langle q, \mathbf{P}, \mathbf{Prov} \rangle$ where $\mathbf{P} = \text{BM25}(q) \cup \text{DPR}(q)$
    \item Generation and end-to-end training instances are pairs of query and target: $\langle q, t \rangle$
\end{itemize}

The first two phases, DPR and generation, are identical to \kgi{}, specifically \kgi{0}. We use the codes from \citet{kgi_emnlp}\footnote{\url{https://github.com/IBM/kgi-slot-filling}}. 

DPR Stage 1 training is the same training used by \citet{dpr}. The triplets of query, positive passage and ``hard negative'' passages from BM25 are put into batches of 128 instances. The positives and hard negatives from other instances form the ``batch negatives'' for each instance. The DPR bi-encoder model gives each query a probability distribution over the positive, hard negative, and batch negatives. The loss is the negative log-likelihood for the positive.  After DPR Stage 1 training the passages from the corpus are indexed with a Hierarchical Navigable Small World (HNSW) \cite{hnsw} using FAISS \cite{faiss}.

Generation training extends the training of the query encoder and trains the BART$_{\text{LARGE}}$ sequence-to-sequence model on the target sequence output. This training is the same as that described by \citet{rag}.

\subsection{Reranking Training}

The next phase, training the reranking in isolation, begins with gathering the initial retrieval results from DPR and BM25 on the training set.  These results are merged and used as training data for the reranker.

In some datasets there are multiple positive passages. Therefore, we use the negative of the summed log-likelihood for the positive passages as the loss function.  The logits given by the reranker are $\mathbf{z_r}$ and the indices for the correct passages (from the ground truth provenance) are $\mathbf{Prov}$.

\begin{align*}
    loss = - \sum_{i \in \mathbf{Prov}} log(softmax(\mathbf{z_r})_i)
\end{align*}
% I discarded out-of-recall instances, but maybe the positive passages should be added
\begin{table*}[t!]
\begin{center}
%\resizebox{0.8\linewidth}{!}{%
\small
% TODO: SEAL~\cite{seal_kilt}
%\bgroup
\renewcommand{\arraystretch}{1.2} 
\begin{tabular}{r|cc|ccccc}
& \multicolumn{6}{c}{\textbf{T-REx}}\small(Slot Filling) \\ \hline
%\textbf{T-REx}\\
 & \textbf{R-Prec}  & \textbf{Recall@5} & \textbf{Accuracy} & \textbf{F1}  & \textbf{KILT-AC} & \textbf{KILT-F1}\\
\hline 
\retwog~(ours)      & \best{80.70} & \best{89.00} & \best{87.68} & \best{89.93} & \best{75.84} & \best{77.05} \\
\kgi{1}~\cite{kgi_emnlp}    & 74.36 & 83.14 & \sbest{84.36} & \sbest{87.24} & \sbest{69.14} & \sbest{70.58} \\
%Old KILT-WEB 2~\cite{kilt_web2}   & 71.86 & \sbest{84.76} & 82.20 & 85.28 & 62.92 & 64.60 \\
KILT-WEB 2~\cite{kilt_web2}   & \sbest{75.64} &	\sbest{87.57} & 81.34 & 84.46 & 64.64 & 66.64 \\
SEAL~\cite{seal_kilt}         & 67.80 & 81.52 & 83.72 & 86.53 & 60.08 & 61.72 \\
\kgi{0}~\cite{kgi_emnlp}      & 59.70 & 70.38 & 77.90 & 81.31 & 55.54 & 56.79 \\
%DensePhrases~\cite{densephrases} & 37.62 & 40.07 & 53.90 & 61.74 & 27.84 & 32.34 \\
% RAG~\cite{kilt}          & 28.68 & 33.04 & 59.20 & 62.96 & 23.12 & 23.94 \\
% GENRE~\cite{genre}        & 79.42 & 85.33 &  0.10 &  7.67 &  0.04 &  6.66 \\
\hline \\

& \multicolumn{6}{c}{\textbf{Natural Questions}}\small(Question Answering) \\ \hline
 & \textbf{R-Prec}  & \textbf{Recall@5} & \textbf{Accuracy} & \textbf{F1}  & \textbf{KILT-AC} & \textbf{KILT-F1}\\
\hline 
\retwog~(ours)           & \best{70.78} & \best{76.63} & \sbest{51.73} & \sbest{60.97} & \best{43.56} & \best{49.80} \\
SEAL~\cite{seal_kilt}    & 63.16 & 68.19 & \best{53.74} & \best{62.24} & \sbest{38.78} & \sbest{44.40} \\
\kgi{0}~\cite{kgi_emnlp} & \sbest{63.71} & 70.17 & 45.22 & 53.38 & 36.36 & 41.83 \\
KILT-WEB 2~\cite{kilt_web2}   & 59.83 & \sbest{71.17} & 51.59 & 60.83 & 35.32 & 40.73 \\
RAG~\cite{kilt}          & 59.49 & 67.06 & 44.39 & 52.35 & 32.69 & 37.91 \\
%BERT+DPR~\cite{kilt}     & \sbest{60.66} & 46.79 & 38.64 & 47.09 & 31.99 & 37.58 \\
%BART+DPR~\cite{kilt}     & 54.29 & 65.52 & 41.27 & 49.54 & 30.06 & 34.72 \\
%MultiDPR~\cite{multidpr} & 59.42 & \sbest{68.24} & 39.75 & 48.43 & 29.09 & 34.70\\
\hline \\

& \multicolumn{6}{c}{\textbf{TriviaQA}}\small(Question Answering) \\ \hline
 & \textbf{R-Prec}  & \textbf{Recall@5} & \textbf{Accuracy} & \textbf{F1}  & \textbf{KILT-AC} & \textbf{KILT-F1}\\
\hline 
\retwog~(ours)           & \best{72.68} & \sbest{74.23} & \best{76.27} & \best{81.40} & \best{57.91} & \best{61.78} \\
SEAL~\cite{seal_kilt}    & \sbest{68.36} & \best{76.36} & 70.86 & 77.29 & \sbest{50.56} & \sbest{54.99} \\
KILT-WEB 2~\cite{kilt_web2}  & 58.85 & 71.55 & \sbest{72.73} & \sbest{79.54} & 45.55 & 49.57 \\
\kgi{0}~\cite{kgi_emnlp} & 60.49 & 63.54 & 60.99 & 66.55 & 42.85 & 46.08 \\
MultiDPR~\cite{multidpr} & 61.49 & 68.33 & 59.60 & 66.53 & 42.36 & 46.19 \\
%RAG~\cite{kilt}          & 48.68 & 57.13 & \sbest{71.27} & \sbest{75.88} & 38.13 & 40.15 \\
%BERT+DPR~\cite{kilt}     & 43.40 & 31.45 & 70.38 & 74.41 & 34.48 & 36.28 \\
%BART+DPR~\cite{kilt}     & 44.49 & 56.99 & 58.55 & 67.79 & 31.40 & 35.34 \\
%KILT-WEB 2 (anonymous)   &  0.00 &  0.00 & 73.06 & 80.33 &  0.00 &  0.00 \\
\hline \\

& \multicolumn{6}{c}{\textbf{FEVER}} \small(Fact Checking) \\ \hline
%\textbf{FEVER}\\
 & \textbf{R-Prec}  & \textbf{Recall@5} & \textbf{Accuracy} &   & \textbf{KILT-AC} & \\
\hline
\retwog~(ours)      & \best{88.92} & \best{92.52} & \best{89.55} &  & \best{78.53} & \\
SEAL~\cite{seal_kilt}    & \sbest{81.45} & \sbest{89.56} & \sbest{89.54} & & \sbest{71.28} & \\
KILT-WEB 2~\cite{kilt_web2}  & 74.77 & 87.89 & 88.99 & & 65.68 & \\
\kgi{0}~\cite{kgi_emnlp}     & 75.60 & 84.95 & 85.58 &  & 64.41 & \\
MultiDPR~\cite{multidpr} & 74.48 & 87.52 & 86.32 & & 63.94 & \\
%RAG~\cite{kilt}         & 61.94 & 75.55 & 86.31 &  & 53.45 & \\
%GENRE~\cite{genre}       & \sbest{83.64} & \sbest{88.15} & 0.00  &  &  0.00 & \\

\hline \\
& \multicolumn{6}{c}{\textbf{Wizard of Wikipedia}} \small(Dialog) \\ \hline
 & \textbf{R-Prec}  & \textbf{Recall@5} & \textbf{Rouge-L} & \textbf{F1}  & \textbf{KILT-RL} & \textbf{KILT-F1}\\
\hline 
Hindsight~\cite{hindsight_arxiv} & 56.08 & 74.27 & \best{17.06} & \best{19.19} & \best{11.92} & \best{13.39} \\
\retwog~(ours)       & \best{60.10} & \best{79.98} & \sbest{16.76} & \sbest{18.90} & \sbest{11.39} & \sbest{12.98} \\
SEAL~\cite{seal_kilt}    & 57.55 & \sbest{78.96} & 16.65 & 18.34 & 10.45 & 11.63 \\
\kgi{0}~\cite{kgi_emnlp}      & 55.37 & 78.45 & 16.36 & 18.57 & 10.36 & 11.79 \\
RAG~\cite{kilt}          & \sbest{57.75} & 74.61 & 11.57 & 13.11 & 7.59 & 8.75 \\
KILT-WEB 2~\cite{kilt_web2}  & 41.54 & 68.25 & 13.94 & 15.66 & 6.55 & 7.57 \\
% MultiDPR~\cite{multidpr} & 41.06 & 67.13 & 13.27 & 15.12 & 5.91 & 6.96 \\
% GENRE~\cite{genre}         & \best{62.88} & 77.74 & 0.00 & 0.00 & 0.00 & 0.00  \\
\end{tabular}
%\egroup
\end{center}
\caption{KILT leaderboard top systems}
\label{tbl.test}
\end{table*}

\subsection{End-to-End Training}

Training end-to-end poses a special challenge. In RAG, the gradient propagates to the query encoder because the inner product between the query vector and the passage vector is used to weight the influence of each sequence, a process RAG calls marginalization. The inputs to the BART model are sequences ($s_j = p_j\ \text{\small [SEP]}\ q$) that comprise a query $q$ plus retrieved passage $p_j$. The probability for each sequence is determined from the softmax over the retrieval (or reranker) scores for the passage. The probability for each target token $t_i$ given the sequence $s_j$ is a softmax over BART's token prediction logits.  The loss therefore is a negative log-likelihood summed over all target tokens and sequences, weighted by each sequence's probability. 

Consider that in \retwog{} the score from the reranker, not the initial retrieval, is used to weight the impact of each sequence in generation.  This allows the reranker to be trained through the ground truth on target output, but it means the gradient for the query encoder will be zero since the marginalization no longer depends on the inner product from the query and passage representation vectors.

\begin{align*}
    P(s_j) & = softmax(\mathbf{z_r})_j\\
    P(t_i | s_j) & = softmax(\text{BART}(s_j)_i)_{t_i} \\
    loss  & = - \sum_{i,j} log \left(P(t_i | s_j) \cdot P(s_j) \right)
\end{align*}

We consider three possible resolutions to this issue.
\begin{itemize}
    \item Combine the DPR and reranker scores
    \item Freeze the query encoder
    
    \item Online Knowledge Distillation
\end{itemize}

The first candidate solution is tempting but fatally flawed.  By adding the log softmax from DPR and the reranker we can ensure that both systems are trained through impact in generation.  However, if the DPR score is added to the reranker score, then the DPR score is being trained to provide a complementary signal to the reranker. Therefore, when DPR is used to gather the candidate passages, it does not give the highest scores to the passages that are most likely to be relevant, but instead gives the highest scores to the passages the reranker is most likely to underrate. We find that this theoretical concern is also a practical concern, as DPR performance (and overall system performance) declines greatly when trained in this way.

The simplest solution is to freeze the parameters of the query encoder, training only the reranker and generation components.  We find this is indeed the best solution for one of our datasets, Wizard of Wikipedia.  Note that DPR has already been trained in two phases, first from the provenance ground truth and then again in generation training in the RAG model. %This approach can be very limiting in cases when ground truth provenance is not available, since the neural IR of DPR can not be trained at all in such a setting. 

The third solution is our novel application of knowledge distillation \cite{knowledgeDistillation}. We use the reranker as a teacher model to provide labels to the DPR student model.  We distill the knowledge across architectures: from an interaction model to a representation model. Further, this knowledge distillation occurs online, while the reranker is being trained. The loss for the initial retrieval is therefore the KL-divergence between the probability distribution it gives over the retrieved passages and the reranker's probability distribution over the same passages. A temperature hyperparameter $T$ smooths these distributions to prevent excessive loss and stabilize training.

\begin{align*}
   loss & = D_{KL}\left({\scriptstyle softmax}\left(\frac{\mathbf{z_s}}{T}\right) \middle\| {\scriptstyle softmax}\left(\frac{\mathbf{z_t}}{T}\right)\right) \cdot T^2 
\end{align*}

The knowledge distillation has the usual advantage of providing signal not only of positive and negative instances, but degrees of negativeness.  In addition, since we retrieve $n = 12$ passages from DPR but only use the top-$k$ ($k = 5$) for generation, the knowledge distillation loss is providing a (soft) label for more passages.

% CONSIDER: mention the use of multiple optimizers?

%\mrglass{This paper clearly states what claims are being investigated.}

%\mrglass{This paper explains how the results substantiate the claims.}

%\mrglass{This paper explicitly identifies limitations or technical assumptions.}

\subsection{Inference}

At inference time the query is encoded using the DPR query encoder and the top-12 passages from the HNSW index are returned. The query is also passed to BM25 search, specifically Anserini\footnote{\url{https://github.com/castorini/anserini}}, gathering the top-12 BM25 results. Both sets of passages are passed to the reranker and scored. The top-5 passages are then joined with the query and passed to BART$_{\text{LARGE}}$ to generate the output. The five output sequences are weighted according to the softmax over the reranker scores to produce the final output.

%\FloatBarrier
\section{Experiments}
We test our model on five datasets, over four distinct tasks in the KILT benchmark: slot filling, question answering, fact checking and dialog. Figure \ref{fig.taskexamples} shows an example of these four tasks.
%\hui{In the following paragraphs, ``dataset'' was used. Should it be changed to ``task'' to be consistent with the term used here and in other places?}

The slot filling dataset, T-REx \cite{trex}, provides as input a head entity and relation, and expects as output the entity or term that fills the slot, also called the tail entity.  
%For example, \texttt{Star Trek [SEP] creator} has the slot filler of \texttt{Gene Roddenberry}.  
The T-REx dataset contains 2.3M instances. %Since this would require many days for training, 
We use only 370k training instances by downsampling the relations that occur more than 5000 times. This reduces the training time required while keeping state-of-the-art performance. The development and test sets each have 5k instances.

The question answering datasets are ``open'' versions of Natural Questions~\cite{naturalquestions} and TriviaQA~\cite{triviaqa}. Unlike the original versions, the relevant Wikipedia page must be found by a retrieval step. The training sets for Natural Questions and TriviaQA contain 87k and 62k questions, with another 3k and 5k for the development and 1.4k and 6.5k for test.
%87372 nq-train-kilt.jsonl
%2837 nq-dev-kilt.jsonl
%1444 nq-test_without_answers-kilt.jsonl
%61844 triviaqa-train-kilt.jsonl
%5359 triviaqa-dev-kilt.jsonl
%6586 triviaqa-test_without_answers-kilt.jsonl

The fact checking dataset in KILT is FEVER (Fact Extraction and VERification). It is a combination of the two FEVER versions \cite{fever,fever2}  omitting the \textsc{NotEnoughInfo} class. There are approximately 10k instances in the development and test sets, and 100k for training. FEVER is a classification task, but we cast it as a generation task by training the model to generate either the token ``SUPPORTS'' or ``REFUTES''.
% 10444 fever-dev-kilt.jsonl
% 104966 fever-train-kilt.jsonl
% 10100 fever-test_without_answers-kilt.jsonl

Wizard of Wikipedia \cite{wizardofwikipedia} is the dialog dataset. The input is a short dialog history ending with the information seeker's turn. The expected output is a fact presented conversationally or just an utterance or question mentioning content from a relevant Wikipedia page.  It is the smallest dataset with approximately 3k instances in development and test and 64k in train.
%\hui{The expected output is not always a fact, sometimes just utterances/questions mentioning some content/entities in the text.}
% 3054 wow-dev-kilt.jsonl
% 63734 wow-train-kilt.jsonl
% 2944 wow-test_without_answers-kilt.jsonl

% TODO: or stacked and single column

\begin{table*}[t!]
\begin{center}
%\resizebox{0.8\linewidth}{!}{%
\small
%\begingroup
\renewcommand{\arraystretch}{1.2} 
\begin{tabular}{r|cc|cc|cc|cc|cc}
& \multicolumn{2}{c|}{\textbf{T-REx}} & \multicolumn{2}{c|}{\textbf{NQ}} &
\multicolumn{2}{c|}{\textbf{TriviaQA}} &
\multicolumn{2}{c|}{\textbf{FEVER}} & \multicolumn{2}{c}{\textbf{WoW}} \\ \hline
%\textbf{T-REx}\\
 & \textbf{R-Prec}  & \textbf{R@5} & \textbf{R-Prec}  & \textbf{R@5}  & \textbf{R-Prec}  & \textbf{R@5}  & \textbf{R-Prec}  & \textbf{R@5}  &
 \textbf{R-Prec}  & \textbf{R@5}\\
\hline 
BM25          & 46.88 & 69.59 
& 24.99 & 42.57 & 26.48 & 45.57
& 42.73 & 70.48 & 27.44 & 45.74 \\
% & 46.88{\small $\pm$1.38} & 69.59{\small $\pm$1.26}
DPR Stage 1   & 49.02 & 63.34
& 56.64 & 64.38 & 60.12 & 64.04
& 75.49 & 84.66 & 34.74 & 60.22 \\
% & 49.02{\small $\pm$1.39} & 63.34{\small $\pm$1.31}
% & 75.49{\small $\pm$0.79} & 84.66{\small $\pm$0.67}
\kgi{0} DPR   & 65.02 & 75.52
& 64.65 & 69.60 & 60.55 & 63.65
& 80.34 & 86.53 & \best{48.04} & \best{71.02} \\ 
% & 65.02{\small $\pm$1.32} & 75.52{\small $\pm$1.16}
% & 80.34{\small $\pm$0.73} & 86.53{\small $\pm$0.63}
\retwog{} DPR & \best{67.16} & \best{76.42} 
& \best{65.88} & \best{70.90} & \best{62.33} & \best{65.72}
& \best{84.13} & \best{87.90} & 47.09 & 69.88 \\
% & 67.16{\small $\pm$1.30} & 76.42{\small $\pm$1.15}
% & 84.13{\small $\pm$0.66} & 87.90{\small $\pm$0.60}
\hline
\kgi{0} DPR+BM25       & 60.48 & 80.06 
& 36.91 & 66.94 & 40.81 & 64.79
& 65.95 & 90.34 & 35.63 & 68.47 \\
% & 60.48{\small $\pm$1.36} & 80.06{\small $\pm$1.08}
Reranker Stage 1       & 81.22 & 87.00 
& 70.78 & 73.05 & \best{71.80} & \best{71.98}
& 87.71 & 92.43 & 55.50 & \best{74.98} \\
% & 81.22{\small $\pm$1.08} & 87.00{\small $\pm$0.89} 

% reranking KGI0 DPR and BM25
%\retwog{} Reranker     & 80.52 & 86.80 
%& 70.53 & 73.42 & 60.59 & 69.43
%& 90.02 & 92.57 & 57.40 & 74.69 \\

% reranker on Re2G DPR and BM25

\retwog{} Reranker     & \best{81.24} & \best{88.58}
& \best{70.92} & \best{74.79} & 60.37 & 70.61
& \best{90.06} & \best{92.91} & \best{57.89} & 74.62 \\

%\retwog{}-KD Reranker  & 80.08 & 86.85 & & & & & 89.97 & 92.48 & 57.99 & 74.49 \\
\end{tabular}
%\endgroup
\end{center}
\caption{Development Set Results for Retrieval}
\label{tbl.retrieve}
\end{table*}

%T-REx: total = 5000, gain = 406, loss = 199; gain prov = 275 vs. 21
%FEVER: total = 10444, gain = 513, loss = 178; gain prov = 343 vs. 33
%WoW:   total = 3054, gain = 1258, loss = 1237; gain prov = 349 vs. 111
% TODO: show the connection between output gain and prov gain in a graph

% TODO: recall gains from BM25 / DPR

For all tasks, systems are expected to produce the target output as well as justify it with provenance information from the KILT knowledge source. The metrics of R-Precision and Recall@5 measure the correctness of the provenance. R-Precision measures what fraction of the $R$ documents in the ground truth provenance ($|\mathbf{Prov}| = R$) are present in the top-$R$ documents returned by the system.  Accuracy and (token-level) F1 measure the correctness of the generated output. For Wizard of Wikipedia, Rouge-L \cite{rouge} is used instead of accuracy, since systems are very unlikely to generate the exact target output.  The metrics of KILT-Accuracy, KILT-F1 and, for Wizard of Wikipedia, KILT-Rouge-L are the underlying metric (e.g. Accuracy) for instances where R-Precision is one, otherwise zero.  These metrics indicate output correctness when provenance is also correctly supplied.

Table \ref{tbl.test} shows the performance of \retwog{} on the KILT leaderboard.  We achieved 9\%, 31\%, 34\%, 22\% and 10\% relative gains over the previous state-of-the-art on the headline KILT metrics for T-REx, Natural Questions, TriviaQA, FEVER, and Wizard of Wikipedia, respectively. Furthermore, \retwog{} has held the lead in the headline KILT metrics in all datasets except for Wizard of Wikipedia where it is now second best. %, \retwog{} is the best on all metrics across all five datasets attempted. 

%The closest competition in retrieval metrics is GENRE. This system, as described in Section~\ref{sec.relatedwork}, uses a Wikipedia-specific approach to retrieval: generating the title of the Wikipedia page as in an entity-linking task. In contrast our system can be applied to any corpus and provides passage-level granularity.

Since our submission to the KILT leaderboard for the Wizard of Wikipedia, a new system called Hindsight~\cite{hindsight_arxiv} achieved even better results on the generation metrics on that particular task. The new system of SEAL has also achieved top results for some metrics on the Natural Questions and TriviaQA benchmarks.
% TODO: describe Hindsight

\subsection{Retrieval}

Table \ref{tbl.retrieve} examines how the retrieval improves through each step of training. In the first half of the table we consider the initial retrieval alone. DPR Stage 1 is the DPR training described earlier - training only from the provenance ground truth with batch negatives and hard negatives from BM25.  \kgi{0} further trains the query encoder of DPR Stage 1 through its impact in generating the target output. Finally \retwog{} extends the training of DPR with online knowledge distillation from the reranker. This step is beneficial in two of the three datasets, while the previous steps improve performance across all datasets.

In the second half of the table we examine the improvement in reranking. The baseline of \kgi{0} DPR+BM25 merges the results of \kgi{0}'s DPR and BM25 by scoring each passage by the sum of the inverse rank from each method. For both T-REx and FEVER, even this simple approach to ensembling DPR and BM25 improves Recall@5, although not R-Precision. 
%We use this set of initial results for the reranking evaluations. 
Following reranker training using the provenance ground truth (Reranker Stage 1), we find improvement over DPR across all five datasets on both retrieval metrics. The reranker's improvement following end-to-end training is mixed. In FEVER and Wizard of Wikipedia there is substantial gain in R-Precision, approximately 2\%.  T-REx and Natural Questions are flat. However, there is a sharp decline in the performance of TriviaQA, in retrieval metrics. This is true despite the fact that retrieving these passages greatly improves answer accuracy and F1.  This suggests some incompleteness in the provenance ground truth for TriviaQA.
%However, Recall@5 remains flat, which is a logical but not inevitable consequence of the fact that the loss is only non-zero for the top-5 passages in end-to-end training. 

%To understand the impact of different components we ran ablations of \retwog{} over each of the five datasets. We considered a variant that eliminates the online knowledge distillation, and a variant that removes results from BM25, using 24 DPR results rather than 12 from both DPR and BM25.

%These variants performed worse in four out of five datasets. Online knowledge distillation failed to improve for Wizard of Wikipedia and ensembling with BM25 failed to improve for Natural Questions.
\subsection{Ablations}
%\section{Ablations}\label{apx.ablation}

\begin{table*}[t!]
\begin{center}
%\resizebox{0.8\linewidth}{!}{%
\small
%\begingroup
\renewcommand{\arraystretch}{1.2} 
\begin{tabular}{r|cc|ccccc}
& \multicolumn{6}{c}{\textbf{T-REx}}~\small(Slot Filling) \\ \hline
%\textbf{T-REx}\\
 & \textbf{R-Prec}  & \textbf{Recall@5} & \textbf{Accuracy} & \textbf{F1}  & \textbf{KILT-AC} & \textbf{KILT-F1}\\
\hline 
\retwog        & 81.24{\small $\pm$1.08} & 88.58{\small $\pm$0.84} & 86.60{\small $\pm$0.94} & 89.20{\small $\pm$0.81} & 75.66{\small $\pm$1.19} & 77.08{\small $\pm$1.15}\\
% & \best{81.24} & 88.58 & 86.60 & 89.20 & 75.66 & \best{77.08} \\ 
              % 81.24%	88.58%	86.60%	89.20%	75.66%	77.08%
\retwog{}-KD   & 81.08{\small $\pm$1.09} & 88.84{\small $\pm$0.83} & 87.00{\small $\pm$0.93} & 89.46{\small $\pm$0.80} & 75.72{\small $\pm$1.19} & 77.00{\small $\pm$1.15}\\
%& 81.08 & \best{88.84} & \best{87.00} & \best{89.46} & \best{75.72} & 77.00 \\ 
               % 81.08%	88.84%	87.00%	89.46%	75.72%	77.00%
\retwog{}-BM25 & 71.92{\small $\pm$1.25} & 78.67{\small $\pm$1.10} & 79.48{\small $\pm$1.12} & 82.52{\small $\pm$1.00} & 66.58{\small $\pm$1.31} & 67.93{\small $\pm$1.28}\\
%& 71.92 & 78.67 & 79.48 & 82.52 & 66.58 & 67.93 \\ 
              % 71.92%	78.67%	79.48%	82.52%	66.58%	67.93%
\kgi{0}       & 65.02{\small $\pm$1.32} & 75.52{\small $\pm$1.16} & 77.52{\small $\pm$1.16} & 80.91{\small $\pm$1.03} & 60.18{\small $\pm$1.36} & 61.38{\small $\pm$1.34}\\
% & 65.02 & 75.52 & 77.52 & 80.91 & 60.18 & 61.38 \\
              % 65.02%	75.52% 77.52%	80.91%	60.18%	61.38%
\hline \\

& \multicolumn{6}{c}{\textbf{Natural Questions}}~\small(Question Answering) \\ \hline
 & \textbf{R-Prec}  & \textbf{Recall@5} & \textbf{Accuracy} & \textbf{F1}  & \textbf{KILT-AC} & \textbf{KILT-F1}\\
\hline 
\retwog        & 70.92{\small $\pm$1.67} & 74.79{\small $\pm$1.27} & 46.70{\small $\pm$1.84} & 62.44{\small $\pm$1.65} & 39.23{\small $\pm$1.80} & 50.90{\small $\pm$1.76}\\
\retwog{}-KD   & 69.72{\small $\pm$1.69} & 73.73{\small $\pm$1.30} & 46.56{\small $\pm$1.84} & 61.68{\small $\pm$1.67} & 38.24{\small $\pm$1.79} & 49.93{\small $\pm$1.76}\\
\retwog{}-BM25 & 70.88{\small $\pm$1.67} & 74.39{\small $\pm$1.28} & 46.70{\small $\pm$1.84} & 61.98{\small $\pm$1.66} & 39.41{\small $\pm$1.80} & 50.91{\small $\pm$1.76}\\
\kgi{0}       & 64.65{\small $\pm$1.76} & 69.60{\small $\pm$1.39} & 40.50{\small $\pm$1.81} & 55.07{\small $\pm$1.71} & 32.96{\small $\pm$1.73} & 42.87{\small $\pm$1.75}\\
\hline \\

& \multicolumn{6}{c}{\textbf{TriviaQA}}~\small(Question Answering) \\ \hline
 & \textbf{R-Prec}  & \textbf{Recall@5} & \textbf{Accuracy} & \textbf{F1}  & \textbf{KILT-AC} & \textbf{KILT-F1}\\
\hline 
\retwog        & 72.01{\small $\pm$1.20} & 73.16{\small $\pm$0.98} & 74.01{\small $\pm$1.17} & 80.86{\small $\pm$0.99} & 56.04{\small $\pm$1.33} & 60.91{\small $\pm$1.27}\\
\retwog{}-KD   & 72.01{\small $\pm$1.20} & 73.16{\small $\pm$0.98} & 73.80{\small $\pm$1.18} & 80.62{\small $\pm$1.00} & 56.04{\small $\pm$1.33} & 60.84{\small $\pm$1.28} \\
\retwog{}-BM25 & 71.10{\small $\pm$1.21} & 68.60{\small $\pm$1.03} & 68.59{\small $\pm$1.24} & 76.68{\small $\pm$1.08} & 52.85{\small $\pm$1.34} & 58.37{\small $\pm$1.29} \\
\kgi{0}       & 61.13{\small $\pm$1.31} & 63.12{\small $\pm$1.08} & 60.68{\small $\pm$1.31} & 66.61{\small $\pm$1.20} & 44.00{\small $\pm$1.33} & 47.35{\small $\pm$1.31}\\
\hline \\

& \multicolumn{6}{c}{\textbf{FEVER}}~\small(Fact Checking) \\ \hline
%\textbf{FEVER}\\
 & \textbf{R-Prec}  & \textbf{Recall@5} & \textbf{Accuracy} &   & \textbf{KILT-AC} & \\
\hline
\retwog       & 90.06{\small $\pm$0.53} & 92.91{\small $\pm$0.47} & 91.05{\small $\pm$0.55} &  & 80.56{\small $\pm$0.76} &\\
%& \best{90.06} & \best{92.91} & \best{91.05} &  & \best{80.56} &  \\
% 90.06%	92.91%	91.05%	80.56%

\retwog{}-KD   & 89.85{\small $\pm$0.54} & 92.48{\small $\pm$0.48} & 90.78{\small $\pm$0.55} &  & 80.14{\small $\pm$0.77} &\\
%& 89.85 & 92.48 & 90.78 &  & 80.14 &  \\
             % 89.85%	92.48%	90.78%	80.14%

\retwog{}-BM25 &88.36{\small $\pm$0.57} & 88.46{\small $\pm$0.59} & 90.63{\small $\pm$0.56} &  & 78.74{\small $\pm$0.78} & \\
%& 88.36 & 88.46 & 90.63 &  & 78.74 &  \\
             % 88.36%	88.46%	90.63%	78.74%

\kgi{0}       &80.34{\small $\pm$0.73} & 86.53{\small $\pm$0.63} & 87.84{\small $\pm$0.63} &  & 70.06{\small $\pm$0.88} &\\
%& 80.34 & 86.53 & 87.84 &  & 70.06 &  \\
             % 80.34%	86.53%	87.84%	70.06%

\hline \\
& \multicolumn{6}{c}{\textbf{Wizard of Wikipedia}}~\small(Dialog) \\ \hline
 & \textbf{R-Prec}  & \textbf{Recall@5} & \textbf{Rouge-L} & \textbf{F1}  & \textbf{KILT-RL} & \textbf{KILT-F1}\\
\hline 

\retwog       & 56.48{\small $\pm$1.76} & 74.00{\small $\pm$1.56} & 17.29{\small $\pm$0.52} & 19.35{\small $\pm$0.57} & 11.37{\small $\pm$0.58} & 12.75{\small $\pm$0.63}\\
% & 56.48 & 74.00 & \best{17.29} & 19.35 & 11.37 & 12.75 \\
 % 56.48%	74.00%	17.29%	19.35%  11.37%	12.75%	

\retwog{}-KD   & 57.89{\small $\pm$1.75} & 74.62{\small $\pm$1.54} & 17.26{\small $\pm$0.52} & 19.39{\small $\pm$0.57} & 11.61{\small $\pm$0.58} & 13.14{\small $\pm$0.64}\\% & \best{57.89} & \best{74.62} & 17.26 & \best{19.39} & \best{11.61} & \best{13.14} \\
              % 57.89%	74.62%	17.26%	19.39%	11.61% 13.14%	

\retwog{}-BM25 & 55.83{\small $\pm$1.76} & 72.72{\small $\pm$1.58} & 17.15{\small $\pm$0.51} & 19.17{\small $\pm$0.56} & 11.13{\small $\pm$0.57} & 12.52{\small $\pm$0.63}\\
%& 55.83 & 72.72 & 17.15 & 19.17 & 11.13 & 12.52 \\
 % 55.83%	72.72%	17.15%	19.17%  11.13%	12.52%	

\kgi{0}      & 48.04{\small $\pm$1.77} & 71.02{\small $\pm$1.61} & 16.75{\small $\pm$0.48} & 19.04{\small $\pm$0.53} & 9.48{\small $\pm$0.53} & 10.74{\small $\pm$0.59}\\  
%& 48.04 & 71.02 & 16.75 & 19.04 & 9.48 & 10.74 \\
 % 48.04%	71.02%	16.75%	19.04%  9.48%   10.74%	

\end{tabular}
%\endgroup
\end{center}
\caption{Development Set Results for \retwog{} Variations}
\label{tbl.dev}
\end{table*}

Table \ref{tbl.dev} explores ablations of the \retwog{} system. The point estimates and 95\% confidence intervals are reported. 
%In bold are the best results or results within the confidence interval of the best result.
\retwog{}-KD excludes the online knowledge distillation, instead freezing the query encoder when training the reranker and generator during end-to-end training.
\retwog{}-BM25 excludes BM25 results, fetching 24 passages from DPR rather than 12 from DPR and 12 from BM25. The passages are still reranked.
\kgi{0} is the baseline system, without a reranker and therefore also without BM25 results or online knowledge distillation during training.

Both online knowledge distillation and ensembling with BM25 improve performance in four out of five datasets. Online knowledge distillation failed to improve for Wizard of Wikipedia and ensembling with BM25 failed to improve for Natural Questions.

%The FEVER and TriviaQA datasets show the simplest pattern where each component: reranking, including BM25 results, and online knowledge distillation all produce gains.%, although these gains do not reach significance for online knowledge distillation. 
%In T-REx and Wizard of Wikipedia the impact of reranking and including BM25 results is still clear, but the online knowledge distillation has mixed and non-significant impact on the metrics.  For FEVER and Wizard of Wikipedia most of the gain comes from including the reranker on DPR results. However, for T-REx, incorporating BM25 produces the largest gain.

\section{Analysis}
\label{sec_analysis}

%More details about all the analysis described below can be found in Appendix \ref{apx.analysis}.

%\subsection{Analysis of gains} 
Since the \retwog{} model differs from the \kgi{} model only in the retrieval phase, we hypothesized that its gains in output quality are driven by its better retrieval quality. To test this hypothesis we considered all cases where the \retwog{} model produces better output than the \kgi{0} model and calculated the fraction of such cases where \retwog{}'s rank for the first correct passage is lower than \kgi{0}'s.

We find that for T-REx, NQ, and FEVER the fractions of output gains that could be attributed to improved retrieval and ranking are 67.73\%, 61.08\% and 66.86\% respectively. While for TriviaQA and Wizard of Wikipedia only 36.86\% and 27.74\% of output improvements were accompanied by improved ranking for the correct passage. It is important to note that in Wizard of Wikipedia, many of these improved outputs have only a small gain in token-level F1.

While much of the gain in output quality is attributable to improved recall, at least a third is not.
This reinforces an observation of \citet{kgi_emnlp}, that models trained with better retrieval can produce better output even when the retrieved passages are equivalent at test time.

\subsection{Slot filling error analysis}
To understand the types of errors \retwog{} makes we sampled 50 instances of the development set of the T-REx dataset where the Accuracy and token-level F1 score was zero.

Interestingly, the most common class of error (33/50)  was due to the incompleteness of the ground truth. Often the head entity is ambiguous (19/50), or the relation has multiple fillers (16/50). As an example, consider the following where there are two \emph{Joe O'Donnell} notable for sports in the passages retrieved, and each played for at least two different teams.
\begin{quote}
\small
    Joe O'Donnell {\small [SEP]} member of sports team\\
\textbf{Target:} Buffalo Bills\\
\textbf{\retwog{}}: Dumbarton F.C.\\
\begin{small}

    $\bullet$ Joe O'Donnell (footballer) / Joe O'Donnell (footballer) Joseph 'Joe' O'Donnell (born 3 March 1961) was a Scottish footballer who played for \textbf{Dumbarton} and Stranraer. 
    
    $\bullet$ Joe O'Donnell (American football) / ... fullback, guard and tackle for the University of Michigan from 1960 to 1963. He also played professional football as a guard and tackle for eight seasons for the \textbf{Buffalo Bills}... 
\end{small}    
\end{quote}

When \retwog{} produces genuine errors it is usually because it has selected some entity as a filler related in a different way (6/17) or it has failed to retrieve the necessary passage (9/17).
% adjusted accuracy: 87.68 + (100-87.68) * 33.0/50

% Gain
%  T-REx:  total = 5000, gain = 406, loss = 199; gain prov = 275 vs. 21
%  Fever:  total = 10444, gain = 513, loss = 178; gain prov = 343 vs. 33
%    WoW:  total = 3054, gain = 1258, loss = 1237; gain prov = 349 vs. 111

%\subsubsection{Generation quality analysis}
%\input{generation_analysis}

\section{Conclusions}
\retwog{} considerably advanced the state-of-the-art across five KILT datasets, and still holds the top position in four of the five.
Relative to previous work, such as RAG or \kgi{}, \retwog{} substantially improves both in retrieval and end-to-end performance on slot filling, question answering, fact checking, and dialog. The reranker alone improves performance and enables the inclusion of multiple sources of initial retrieval. This architecture permits us to integrate results from BM25, further improving accuracy.
Our online knowledge distillation is able to improve the performance of DPR in four of the five datasets, despite the loss in end-to-end training not depending on the DPR scores.
Similarly, the ensembling of DPR and BM25, which is enabled by our incorporation of a reranker, benefits four of the five datasets tested.
We have directed our efforts towards improving the retrieval of relevant knowledge.  This also enables improvement in end-to-end performance by supplying better passages to the generation component.
%The impact of \retwog{} on more tasks that require access to external data sources, e.g. open domain question answering, is a promising future direction worth exploring.
%We are also continuing to explore better ways to train retrieval, reranking and generation jointly.
%As shown in the context of slot filling~\cite{kgi_emnlp}, retrieval-based language models allow zero-shot domain adaptation capabilities by replacing the open domain corpus with domain specific document collections. 

Further experiments on domain adaptation of \retwog{} on tasks like question answering or dialog might provide useful insight on the application of this technology to real world use cases.
We are releasing our source code as open source (Apache 2.0 license) to enable further research.
%Our ultimate goal is to use \retwog{} for zero shot domain adaptation, replacing the open domain corpus with domain specific document collections. This will enable us to apply this technology to real world use cases, such as conversational QA that works on user provided data without requiring any domain adaptation effort.

% future work: Re2G pretraining on GPT-3 level corpus

%\FloatBarrier
% Use \bibliography{yourbibfile} instead or the References section will not appear in your paper
\bibliographystyle{plainnat} 
\bibliography{rerankrag}

\begin{thebibliography}{41}
\providecommand{\natexlab}[1]{#1}
\providecommand{\url}[1]{\texttt{#1}}
\expandafter\ifx\csname urlstyle\endcsname\relax
  \providecommand{\doi}[1]{doi: #1}\else
  \providecommand{\doi}{doi: \begingroup \urlstyle{rm}\Url}\fi

\bibitem[Bevilacqua et~al.(2022)Bevilacqua, Ottaviano, Lewis, tau Yih, Riedel,
  and Petroni]{seal_kilt}
Michele Bevilacqua, Giuseppe Ottaviano, Patrick Lewis, Wen tau Yih, Sebastian
  Riedel, and Fabio Petroni.
\newblock Autoregressive search engines: Generating substrings as document
  identifiers.
\newblock \emph{ArXiv}, abs/2204.10628, 2022.

\bibitem[Brown et~al.(2020)Brown, Mann, Ryder, Subbiah, Kaplan, Dhariwal,
  Neelakantan, Shyam, Sastry, Askell, Agarwal, Herbert{-}Voss, Krueger,
  Henighan, Child, Ramesh, Ziegler, Wu, Winter, Hesse, Chen, Sigler, Litwin,
  Gray, Chess, Clark, Berner, McCandlish, Radford, Sutskever, and Amodei]{gpt3}
Tom~B. Brown, Benjamin Mann, Nick Ryder, Melanie Subbiah, Jared Kaplan,
  Prafulla Dhariwal, Arvind Neelakantan, Pranav Shyam, Girish Sastry, Amanda
  Askell, Sandhini Agarwal, Ariel Herbert{-}Voss, Gretchen Krueger, Tom
  Henighan, Rewon Child, Aditya Ramesh, Daniel~M. Ziegler, Jeffrey Wu, Clemens
  Winter, Christopher Hesse, Mark Chen, Eric Sigler, Mateusz Litwin, Scott
  Gray, Benjamin Chess, Jack Clark, Christopher Berner, Sam McCandlish, Alec
  Radford, Ilya Sutskever, and Dario Amodei.
\newblock Language models are few-shot learners.
\newblock In \emph{NeurIPS}, 2020.

\bibitem[Cao et~al.(2021)Cao, Izacard, Riedel, and Petroni]{genre}
Nicola~De Cao, Gautier Izacard, Sebastian Riedel, and Fabio Petroni.
\newblock Autoregressive entity retrieval.
\newblock In \emph{International Conference on Learning Representations}.
  OpenReview.net, 2021.
\newblock URL \url{https://openreview.net/forum?id=5k8F6UU39V}.

\bibitem[Devlin et~al.(2019)Devlin, Chang, Lee, and Toutanova]{bert}
Jacob Devlin, Ming-Wei Chang, Kenton Lee, and Kristina Toutanova.
\newblock {BERT}: Pre-training of deep bidirectional transformers for language
  understanding.
\newblock In \emph{Proceedings of the 2019 Conference of the North {A}merican
  Chapter of the Association for Computational Linguistics: Human Language
  Technologies, Volume 1 (Long and Short Papers)}, pages 4171--4186,
  Minneapolis, Minnesota, June 2019. Association for Computational Linguistics.
\newblock \doi{10.18653/v1/N19-1423}.
\newblock URL \url{https://aclanthology.org/N19-1423}.

\bibitem[Dinan et~al.(2018)Dinan, Roller, Shuster, Fan, Auli, and
  Weston]{wizardofwikipedia}
Emily Dinan, Stephen Roller, Kurt Shuster, Angela Fan, Michael Auli, and Jason
  Weston.
\newblock Wizard of wikipedia: Knowledge-powered conversational agents.
\newblock In \emph{International Conference on Learning Representations}, 2018.

\bibitem[Dinan et~al.(2019)Dinan, Roller, Shuster, Fan, Auli, and
  Weston]{DBLP:conf/iclr/DinanRSFAW19}
Emily Dinan, Stephen Roller, Kurt Shuster, Angela Fan, Michael Auli, and Jason
  Weston.
\newblock Wizard of wikipedia: Knowledge-powered conversational agents.
\newblock In \emph{{ICLR} (Poster)}. OpenReview.net, 2019.

\bibitem[Elsahar et~al.(2018)Elsahar, Vougiouklis, Remaci, Gravier, Hare,
  Laforest, and Simperl]{trex}
Hady Elsahar, Pavlos Vougiouklis, Arslen Remaci, Christophe Gravier, Jonathon
  Hare, Frederique Laforest, and Elena Simperl.
\newblock {T}-{RE}x: A large scale alignment of natural language with knowledge
  base triples.
\newblock In \emph{Proceedings of the Eleventh International Conference on
  Language Resources and Evaluation ({LREC} 2018)}, Miyazaki, Japan, May 2018.
  European Language Resources Association (ELRA).
\newblock URL \url{https://aclanthology.org/L18-1544}.

\bibitem[Ferragina and Manzini(2000)]{FMIndex}
P.~Ferragina and G.~Manzini.
\newblock Opportunistic data structures with applications.
\newblock In \emph{Proceedings 41st Annual Symposium on Foundations of Computer
  Science}, pages 390--398, 2000.
\newblock \doi{10.1109/SFCS.2000.892127}.

\bibitem[Glass et~al.(2021)Glass, Rossiello, Chowdhury, and Gliozzo]{kgi_emnlp}
Michael Glass, Gaetano Rossiello, Md~Faisal~Mahbub Chowdhury, and Alfio
  Gliozzo.
\newblock Robust retrieval augmented generation for zero-shot slot filling.
\newblock In \emph{Proceedings of the 2021 Conference on Empirical Methods in
  Natural Language Processing}, pages 1939--1949, Online and Punta Cana,
  Dominican Republic, November 2021. Association for Computational Linguistics.
\newblock \doi{10.18653/v1/2021.emnlp-main.148}.
\newblock URL \url{https://aclanthology.org/2021.emnlp-main.148}.

\bibitem[Guu et~al.(2020)Guu, Lee, Tung, Pasupat, and Chang]{realm}
Kelvin Guu, Kenton Lee, Zora Tung, Panupong Pasupat, and Ming-Wei Chang.
\newblock Realm: Retrieval-augmented language model pre-training.
\newblock \emph{arXiv preprint arXiv:2002.08909}, 2020.

\bibitem[Hinton et~al.(2015)Hinton, Vinyals, and Dean]{knowledgeDistillation}
Geoffrey Hinton, Oriol Vinyals, and Jeffrey Dean.
\newblock Distilling the knowledge in a neural network.
\newblock In \emph{NIPS Deep Learning and Representation Learning Workshop},
  2015.
\newblock URL \url{http://arxiv.org/abs/1503.02531}.

\bibitem[Izacard and Grave(2021)]{fusionInDecoder}
Gautier Izacard and Edouard Grave.
\newblock Leveraging passage retrieval with generative models for open domain
  question answering.
\newblock In \emph{Proceedings of the 16th Conference of the European Chapter
  of the Association for Computational Linguistics: Main Volume}, pages
  874--880, Online, April 2021. Association for Computational Linguistics.
\newblock \doi{10.18653/v1/2021.eacl-main.74}.
\newblock URL \url{https://aclanthology.org/2021.eacl-main.74}.

\bibitem[Johnson et~al.(2017)Johnson, Douze, and J{\'e}gou]{faiss}
Jeff Johnson, Matthijs Douze, and Herv{\'e} J{\'e}gou.
\newblock Billion-scale similarity search with gpus.
\newblock \emph{arXiv preprint arXiv:1702.08734}, 2017.

\bibitem[Joshi et~al.(2017)Joshi, Choi, Weld, and Zettlemoyer]{triviaqa}
Mandar Joshi, Eunsol Choi, Daniel Weld, and Luke Zettlemoyer.
\newblock {T}rivia{QA}: A large scale distantly supervised challenge dataset
  for reading comprehension.
\newblock In \emph{Proceedings of the 55th Annual Meeting of the Association
  for Computational Linguistics (Volume 1: Long Papers)}, pages 1601--1611,
  Vancouver, Canada, July 2017. Association for Computational Linguistics.
\newblock \doi{10.18653/v1/P17-1147}.
\newblock URL \url{https://aclanthology.org/P17-1147}.

\bibitem[Joshi et~al.(2020)Joshi, Chen, Liu, Weld, Zettlemoyer, and
  Levy]{spanbert}
Mandar Joshi, Danqi Chen, Yinhan Liu, Daniel~S. Weld, Luke Zettlemoyer, and
  Omer Levy.
\newblock {S}pan{BERT}: Improving pre-training by representing and predicting
  spans.
\newblock \emph{Transactions of the Association for Computational Linguistics},
  8:\penalty0 64--77, 2020.
\newblock \doi{10.1162/tacl_a_00300}.
\newblock URL \url{https://aclanthology.org/2020.tacl-1.5}.

\bibitem[Karpukhin et~al.(2020)Karpukhin, Oguz, Min, Lewis, Wu, Edunov, Chen,
  and Yih]{dpr}
Vladimir Karpukhin, Barlas Oguz, Sewon Min, Patrick Lewis, Ledell Wu, Sergey
  Edunov, Danqi Chen, and Wen-tau Yih.
\newblock Dense passage retrieval for open-domain question answering.
\newblock In \emph{Proceedings of the 2020 Conference on Empirical Methods in
  Natural Language Processing (EMNLP)}, pages 6769--6781, Online, November
  2020. Association for Computational Linguistics.
\newblock \doi{10.18653/v1/2020.emnlp-main.550}.
\newblock URL \url{https://aclanthology.org/2020.emnlp-main.550}.

\bibitem[Kwiatkowski et~al.(2019)Kwiatkowski, Palomaki, Redfield, Collins,
  Parikh, Alberti, Epstein, Polosukhin, Devlin, Lee, Toutanova, Jones, Kelcey,
  Chang, Dai, Uszkoreit, Le, and Petrov]{naturalquestions}
Tom Kwiatkowski, Jennimaria Palomaki, Olivia Redfield, Michael Collins, Ankur
  Parikh, Chris Alberti, Danielle Epstein, Illia Polosukhin, Jacob Devlin,
  Kenton Lee, Kristina Toutanova, Llion Jones, Matthew Kelcey, Ming-Wei Chang,
  Andrew~M. Dai, Jakob Uszkoreit, Quoc Le, and Slav Petrov.
\newblock Natural questions: A benchmark for question answering research.
\newblock \emph{Transactions of the Association for Computational Linguistics},
  7:\penalty0 452--466, March 2019.
\newblock \doi{10.1162/tacl_a_00276}.
\newblock URL \url{https://aclanthology.org/Q19-1026}.

\bibitem[Lee et~al.(2021)Lee, Sung, Kang, and Chen]{densephrases}
Jinhyuk Lee, Mujeen Sung, Jaewoo Kang, and Danqi Chen.
\newblock Learning dense representations of phrases at scale.
\newblock In \emph{Proceedings of the 59th Annual Meeting of the Association
  for Computational Linguistics and the 11th International Joint Conference on
  Natural Language Processing (Volume 1: Long Papers)}, pages 6634--6647,
  Online, August 2021. Association for Computational Linguistics.
\newblock \doi{10.18653/v1/2021.acl-long.518}.
\newblock URL \url{https://aclanthology.org/2021.acl-long.518}.

\bibitem[Levy et~al.(2017)Levy, Seo, Choi, and Zettlemoyer]{zsre}
Omer Levy, Minjoon Seo, Eunsol Choi, and Luke Zettlemoyer.
\newblock Zero-shot relation extraction via reading comprehension.
\newblock In \emph{Proceedings of the 21st Conference on Computational Natural
  Language Learning ({C}o{NLL} 2017)}, pages 333--342, Vancouver, Canada,
  August 2017. Association for Computational Linguistics.
\newblock \doi{10.18653/v1/K17-1034}.
\newblock URL \url{https://aclanthology.org/K17-1034}.

\bibitem[Lewis et~al.(2020{\natexlab{a}})Lewis, Liu, Goyal, Ghazvininejad,
  Mohamed, Levy, Stoyanov, and Zettlemoyer]{bart}
Mike Lewis, Yinhan Liu, Naman Goyal, Marjan Ghazvininejad, Abdelrahman Mohamed,
  Omer Levy, Veselin Stoyanov, and Luke Zettlemoyer.
\newblock {BART}: Denoising sequence-to-sequence pre-training for natural
  language generation, translation, and comprehension.
\newblock In \emph{Proceedings of the 58th Annual Meeting of the Association
  for Computational Linguistics}, pages 7871--7880, Online, July
  2020{\natexlab{a}}. Association for Computational Linguistics.
\newblock \doi{10.18653/v1/2020.acl-main.703}.
\newblock URL \url{https://aclanthology.org/2020.acl-main.703}.

\bibitem[Lewis et~al.(2020{\natexlab{b}})Lewis, Perez, Piktus, Petroni,
  Karpukhin, Goyal, K\"{u}ttler, Lewis, Yih, Rockt\"{a}schel, Riedel, and
  Kiela]{rag}
Patrick Lewis, Ethan Perez, Aleksandra Piktus, Fabio Petroni, Vladimir
  Karpukhin, Naman Goyal, Heinrich K\"{u}ttler, Mike Lewis, Wen-tau Yih, Tim
  Rockt\"{a}schel, Sebastian Riedel, and Douwe Kiela.
\newblock Retrieval-augmented generation for knowledge-intensive nlp tasks.
\newblock In H.~Larochelle, M.~Ranzato, R.~Hadsell, M.~F. Balcan, and H.~Lin,
  editors, \emph{Advances in Neural Information Processing Systems}, volume~33,
  pages 9459--9474. Curran Associates, Inc., 2020{\natexlab{b}}.

\bibitem[Lin(2004)]{rouge}
Chin-Yew Lin.
\newblock Rouge: A package for automatic evaluation of summaries.
\newblock In \emph{Text summarization branches out}, pages 74--81, 2004.

\bibitem[Liu(2009)]{learningToRank}
Tie-Yan Liu.
\newblock Learning to rank for information retrieval.
\newblock \emph{Information Retrieval}, 3\penalty0 (3):\penalty0 225--331,
  2009.

\bibitem[Maillard et~al.(2021)Maillard, Karpukhin, Petroni, Yih, Oguz,
  Stoyanov, and Ghosh]{multidpr}
Jean Maillard, Vladimir Karpukhin, Fabio Petroni, Wen{-}tau Yih, Barlas Oguz,
  Veselin Stoyanov, and Gargi Ghosh.
\newblock Multi-task retrieval for knowledge-intensive tasks.
\newblock In \emph{{ACL/IJCNLP} {(1)}}, pages 1098--1111. Association for
  Computational Linguistics, 2021.

\bibitem[Malkov and Yashunin(2018)]{hnsw}
Yu~A Malkov and Dmitry~A Yashunin.
\newblock Efficient and robust approximate nearest neighbor search using
  hierarchical navigable small world graphs.
\newblock \emph{IEEE transactions on pattern analysis and machine
  intelligence}, 42\penalty0 (4):\penalty0 824--836, 2018.

\bibitem[Nguyen et~al.(2016)Nguyen, Rosenberg, Song, Gao, Tiwary, Majumder, and
  Deng]{msmarco}
Tri Nguyen, Mir Rosenberg, Xia Song, Jianfeng Gao, Saurabh Tiwary, Rangan
  Majumder, and Li~Deng.
\newblock Ms marco: A human generated machine reading comprehension dataset.
\newblock In \emph{CoCo@ NIPS}, 2016.

\bibitem[Nogueira and Cho(2019)]{rerankBERT}
Rodrigo Nogueira and Kyunghyun Cho.
\newblock Passage re-ranking with bert.
\newblock \emph{arXiv preprint arXiv:1901.04085}, 2019.

\bibitem[Nogueira et~al.(2020)Nogueira, Jiang, Pradeep, and Lin]{rerankT5}
Rodrigo Nogueira, Zhiying Jiang, Ronak Pradeep, and Jimmy Lin.
\newblock Document ranking with a pretrained sequence-to-sequence model.
\newblock In \emph{Findings of the Association for Computational Linguistics:
  EMNLP 2020}, pages 708--718, Online, November 2020. Association for
  Computational Linguistics.
\newblock \doi{10.18653/v1/2020.findings-emnlp.63}.
\newblock URL \url{https://aclanthology.org/2020.findings-emnlp.63}.

\bibitem[Paranjape et~al.(2021)Paranjape, Khattab, Potts, Zaharia, and
  Manning]{hindsight_arxiv}
Ashwin Paranjape, Omar Khattab, Christopher Potts, Matei Zaharia, and
  Christopher~D Manning.
\newblock Hindsight: Posterior-guided training of retrievers for improved
  open-ended generation.
\newblock \emph{arXiv preprint arXiv:2110.07752}, 2021.

\bibitem[Petroni et~al.(2021)Petroni, Piktus, Fan, Lewis, Yazdani, De~Cao,
  Thorne, Jernite, Karpukhin, Maillard, Plachouras, Rockt{\"a}schel, and
  Riedel]{kilt}
Fabio Petroni, Aleksandra Piktus, Angela Fan, Patrick Lewis, Majid Yazdani,
  Nicola De~Cao, James Thorne, Yacine Jernite, Vladimir Karpukhin, Jean
  Maillard, Vassilis Plachouras, Tim Rockt{\"a}schel, and Sebastian Riedel.
\newblock {KILT}: a benchmark for knowledge intensive language tasks.
\newblock In \emph{Proceedings of the 2021 Conference of the North American
  Chapter of the Association for Computational Linguistics: Human Language
  Technologies}, pages 2523--2544, Online, June 2021. Association for
  Computational Linguistics.
\newblock \doi{10.18653/v1/2021.naacl-main.200}.
\newblock URL \url{https://aclanthology.org/2021.naacl-main.200}.

\bibitem[Piktus et~al.(2021)Piktus, Petroni, Karpukhin, Okhonko, Broscheit,
  Izacard, Lewis, O{\u{g}}uz, Grave, Yih, et~al.]{kilt_web2}
Aleksandra Piktus, Fabio Petroni, Vladimir Karpukhin, Dmytro Okhonko, Samuel
  Broscheit, Gautier Izacard, Patrick Lewis, Barlas O{\u{g}}uz, Edouard Grave,
  Wen-tau Yih, et~al.
\newblock The web is your oyster--knowledge-intensive nlp against a very large
  web corpus.
\newblock \emph{arXiv preprint arXiv:2112.09924}, 2021.

\bibitem[Raffel et~al.(2020)Raffel, Shazeer, Roberts, Lee, Narang, Matena,
  Zhou, Li, and Liu]{t5}
Colin Raffel, Noam Shazeer, Adam Roberts, Katherine Lee, Sharan Narang, Michael
  Matena, Yanqi Zhou, Wei Li, and Peter~J. Liu.
\newblock Exploring the limits of transfer learning with a unified text-to-text
  transformer, 2020.

\bibitem[Robertson and Zaragoza(2009)]{bm25}
Stephen Robertson and Hugo Zaragoza.
\newblock The probabilistic relevance framework: Bm25 and beyond.
\newblock \emph{Found. Trends Inf. Retr.}, 3\penalty0 (4):\penalty0 333--389,
  April 2009.
\newblock ISSN 1554-0669.
\newblock \doi{10.1561/1500000019}.
\newblock URL \url{http://dx.doi.org/10.1561/1500000019}.

\bibitem[Thienes and Pertschuk(2019)]{nboostRerank}
Cole Thienes and Jack Pertschuk.
\newblock Nboost: Neural boosting search results.
\newblock \url{https://github.com/koursaros-ai/nboost}, 2019.

\bibitem[Thorne et~al.(2018{\natexlab{a}})Thorne, Vlachos, Christodoulopoulos,
  and Mittal]{DBLP:conf/naacl/ThorneVCM18}
James Thorne, Andreas Vlachos, Christos Christodoulopoulos, and Arpit Mittal.
\newblock {FEVER:} a large-scale dataset for fact extraction and verification.
\newblock In \emph{{NAACL-HLT}}, pages 809--819. Association for Computational
  Linguistics, 2018{\natexlab{a}}.

\bibitem[Thorne et~al.(2018{\natexlab{b}})Thorne, Vlachos, Christodoulopoulos,
  and Mittal]{fever}
James Thorne, Andreas Vlachos, Christos Christodoulopoulos, and Arpit Mittal.
\newblock Fever: a large-scale dataset for fact extraction and verification.
\newblock In \emph{Proceedings of the 2018 Conference of the North American
  Chapter of the Association for Computational Linguistics: Human Language
  Technologies, Volume 1 (Long Papers)}, pages 809--819, 2018{\natexlab{b}}.

\bibitem[Thorne et~al.(2018{\natexlab{c}})Thorne, Vlachos, Cocarascu,
  Christodoulopoulos, and Mittal]{DBLP:journals/corr/abs-1811-10971}
James Thorne, Andreas Vlachos, Oana Cocarascu, Christos Christodoulopoulos, and
  Arpit Mittal.
\newblock The fact extraction and verification {(FEVER)} shared task.
\newblock \emph{CoRR}, abs/1811.10971, 2018{\natexlab{c}}.

\bibitem[Thorne et~al.(2019)Thorne, Vlachos, Cocarascu, Christodoulopoulos, and
  Mittal]{fever2}
James Thorne, Andreas Vlachos, Oana Cocarascu, Christos Christodoulopoulos, and
  Arpit Mittal.
\newblock The fever2. 0 shared task.
\newblock In \emph{Proceedings of the Second Workshop on Fact Extraction and
  VERification (FEVER)}, pages 1--6, 2019.

\bibitem[Wang et~al.(2011)Wang, Lin, and Metzler]{wang2011cascade}
Lidan Wang, Jimmy Lin, and Donald Metzler.
\newblock A cascade ranking model for efficient ranked retrieval.
\newblock In \emph{Proceedings of the 34th international ACM SIGIR conference
  on Research and development in Information Retrieval}, pages 105--114, 2011.

\bibitem[Wenzek et~al.(2020)Wenzek, Lachaux, Conneau, Chaudhary, Guzm{\'a}n,
  Joulin, and Grave]{ccnet}
Guillaume Wenzek, Marie-Anne Lachaux, Alexis Conneau, Vishrav Chaudhary,
  Francisco Guzm{\'a}n, Armand Joulin, and Edouard Grave.
\newblock {CCN}et: Extracting high quality monolingual datasets from web crawl
  data.
\newblock In \emph{Proceedings of the 12th Language Resources and Evaluation
  Conference}, pages 4003--4012, Marseille, France, May 2020. European Language
  Resources Association.
\newblock ISBN 979-10-95546-34-4.
\newblock URL \url{https://aclanthology.org/2020.lrec-1.494}.

\bibitem[Wu et~al.(2020)Wu, Petroni, Josifoski, Riedel, and Zettlemoyer]{blink}
Ledell Wu, Fabio Petroni, Martin Josifoski, Sebastian Riedel, and Luke
  Zettlemoyer.
\newblock Scalable zero-shot entity linking with dense entity retrieval.
\newblock In \emph{Proceedings of the 2020 Conference on Empirical Methods in
  Natural Language Processing (EMNLP)}, pages 6397--6407, Online, November
  2020. Association for Computational Linguistics.
\newblock \doi{10.18653/v1/2020.emnlp-main.519}.
\newblock URL \url{https://aclanthology.org/2020.emnlp-main.519}.

\end{thebibliography}

\newcommand{\tabitem}{~~\llap{\textbullet}~~}

\section*{Appendix}

\appendix

\section{Hyperparameters}\label{apx.hypers}

We have not done hyperparameter tuning for DPR Stage 1, Generation, or Reranking training. Instead we used hyperparameters similar to the original works on training DPR, BERT reranking and RAG. Table \ref{tbl.hypers} shows the hyperparameters used in our experiments. 

% \mrglass{This paper lists all final (hyper-)parameters used for each model/algorithm in the paper’s experiments.}

\begin{table*}[tbh!]
\begin{center}
\begin{tabular}{rrrr}

\textbf{Hyperparameter} & \textbf{DPR} & \textbf{Reranker}  & \textbf{Generation} \\
\hline
learn rate  & 5e-5 & 3e-5 & 3e-5 \\
batch size & 128 & 32 & 128 \\
epochs & 2 & 1 & 1* \\
warmup instances & 0 & 10\% & 10\% \\
learning schedule & linear & triangular & triangular \\
max grad norm & 1 & 1 & 1 \\
weight decay & 0 & 0 & 0 \\
Adam epsilon & 1e-8 & 1e-8 & 1e-8
\end{tabular}
\end{center}
\caption{\retwog{} hyperparameters}
\label{tbl.hypers}
\end{table*}

For knowledge distillation we used the same hyperparameter settings as Generation. For the additional hyperparameters in online knowledge distillation: temperature and KD learn rate scaling, we experimented with temperatures of 10 and 40 and KD learn rate scaling of 1.0 and 0.1. For our reported results we used a temperature of $10.0$ and a learn rate scaling of $1.0$.

When training using online knowledge distillation, there is a separate optimizer for the query encoder while training generation. This optimizer uses the same hyperparameter settings.

\begin{table}[bh!]
\begin{center}
\begin{tabular}{rr}
\textbf{Hyperparameter} & \textbf{Value} \\
\hline
type & IndexHNSWSQ \\
m & 128 \\
ef search & 128 \\
ef construction & 200 \\
index batch size & 100000 \\
scalar quantizer & 8 
\end{tabular}
\end{center}
\caption{FAISS index hyperparameters}
\label{tbl.faiss_hypers}
\end{table}

%The T-REx dataset contains 2.3M instances.  Since this would require many days for training, we use only 370k instances.

%Our training builds on existing models: $KGI_0$

% \mrglass{This paper states the number and range of values tried per (hyper-)parameter during development of the paper, along with the criterion used for selecting the final parameter setting.}

\begin{table}[bh!]
\begin{center}
\begin{tabular}{rr}
\textbf{Hyperparameter} & \textbf{Value} \\
\hline
DPR passages & 12 \\
BM25 passages & 12 \\
BART sequences & 5 \\
BART beam size & 6 \\
BART length penalty & 1.0 \\
BART minimum length & 2 \\
BART maximum length & 64 
\end{tabular}
\end{center}
\caption{Inference hyperparameters}
\label{tbl.inference_hypers}
\end{table}

Table \ref{tbl.inference_hypers} shows the settings for retrieval and generation used for all datasets.

All results are from a single run. The random seed for python, numpy and pytorch was 42. 

% number of hyperparameters tried

\section{Software Details}\label{apx.software}
% operating system; names and versions of relevant software libraries and frameworks.
We used the following software versions:
\begin{itemize}
\item Ubuntu 18
\item Pytorch 1.7
\item Transformers 4.3.2
\item Anserini 0.4.1\\{\small (commit\\ 3a60106fdc83473d147218d78ae7dca7c3b6d47c)}
\end{itemize}

\section{Model Details}\label{apx.model}

\paragraph{Number of parameters}
\retwog{} uses three BERT$_{BASE}$ transformers: query encoder, passage encoder and reranker. Each has $110M$ parameters. The generation component is a BART$_{LARGE}$ model with $400M$ parameters. There are $730M$ parameters in total.

\paragraph{Computing infrastructure}
Using a single NVIDIA V100 GPU DPR training of two epochs takes approximately 24 hours for T-REx and less than 12 hours for FEVER and WoW. 

Using a two NVIDIA P100 GPUs generation training for 370k T-REx instances takes two days, while FEVER and WoW training completes in half a day.

The FAISS index on the KILT knowledge source requires a machine with large memory, we use machines with 128GB of memory.

\section{Generation Analysis}\label{apx.analysis}
 %We randomly selected 2 separate sets for different \emph{coherence tests}. Each set contains 20 development instances with F1 score in the bottom quintile from Wizard of Wikipedia (WoW). 

%First, we showed one set to the human evaluator where each of the 20 instances was coupled with 3 output texts (output of baseline \kgi{0} DPR+BM25, output of \retwog{} and the target text in ground-truth). 
We examined 20 instances coupled with 3 output texts: the baseline \kgi{0}, \retwog{}, and the target text in the ground-truth. 
The three output texts were presented unlabeled and in random order to avoid bias. %to make sure the evaluator cannot guess which text is from which source. 
For each instance, we read the conversation history and then mark each text either \texttt{\small GOOD}, \texttt{\small OK} or \texttt{\small INCONSISTENT} generation. To our surprise, 5/20 ground-truth target texts are \texttt{\small INCONSISTENT} which indicates the WoW benchmark might have limitations in annotation quality. Both the systems have similar results {\small (\texttt{GOOD/OK/INCONSISTENT} - \retwog{}: 8/2/10; \kgi{0}: 9/2/9)}.
%Target vs. Re2G
%  Target > Re2G: 10
%  Re2G > Target: 4
%  Re2G ~= Target: 6
% \hui{For dialogue evaluation, some common human eval metrics are fluency, coherence besides factual correctness, etc.. Maybe mention here the evaluation is for coherence (wizard response being relevant or not to the conversation)?}

%Second, we asked the evaluator to check the other set of 20 WoW instances but this time pairing \retwog{} generated text with the passages retrieved.
Second, we checked a set of 20 WoW instances where \retwog{}'s F1 score was in the bottom quintile. 
The conversation history was presented along with \retwog{} generated text and the passages retrieved.
%In this particular manual evaluation, he read the conversation history and the target text, then checked the generated text for coherence. 
%He also checked in case of inconsistent generated text, whether it  was due to deficiency in the appropriate passage retrieval.  
% \hui{Is the evaluation here for whether the retrieved passage is relevant to the conversation, or for whether the generated text is consistent given the retrieved passage?}
Manual examination showed 8/20 as \texttt{\small INCONSISTENT} and in 4/8 cases supporting ground-truth passages were not retrieved. 
%\hui{Does this mean that out of the 8 cases where the retrieved passage is not relevant, 4 of them are actually ground truth passages?}
Below is one of the 12/20 cases where \retwog{} generated text was found \texttt{\small CONSISTENT} with respect to the conversation history, although it has low F1 and Rouge-L scores. 

\begin{quote}
\small
\textbf{Conversation History:}
\begin{itemize}
    \item My favorite color is red.
    \item Red is at the end of the spectrum of light, its with orange and opposite of violet.
    \item I didn't know that. What else do you know about red?
\end{itemize}
\textbf{Target:} It's actually a primary color for the RGB and CMYK color model.\\
\textbf{\retwog{}}: It has a dominant wavelength of approximately 625-740 nanometres.
\end{quote}

\begin{table*}[t]
\centering
\begin{tabular}{p{15cm}}
\hline
\textbf{Conversation history:} \\
\tabitem Have you ever been to Niagra Falls? I want to go there, but I know it's very far. Those beautiful 3 waterfalls are right between New York and Ontario, thousands of miles from me! \\
\tabitem I never have but I would love to. It looks beautiful.\\
\tabitem  Yes, they do! They are different sized waterfalls, named Horseshoe, American and Bridal Veil Falls.\\
\tabitem I had no idea. I've never heard that before. Very cool.\\
\tabitem  Yes, and Horseshoe Falls in actually the most powerful waterfall in all of the US!\\
\tabitem I never new any details about it. I have just seen it in pictures and on TV and it looks very neat. \\~\\
\textbf{\retwog{}:}  Horseshoe falls, also known as Canadian Falls, is the largest of the three waterfalls that collectively form Niagara Falls on the Niagara River along with American Falls and Bridal Veil Falls. \\
\texttt{[GOOD]} \\\hline
\textbf{\kgi{0}:} Horseshoe falls is also known as Canadian Falls, which straddles the international border between Canada and the U.S.  \\\texttt{[GOOD]}\\\hline
\textbf{Ground truth:} It's really quite interesting and beautiful. There's over six million cubic feet of water that crosses the fall's crest every 60 seconds. Can you imagine?\\ \texttt{[INCONSISTENT - 6M cubic feet of water crosses for the 3 waterfalls combined, not just Horseshoe falls.]}\\
\hline
\hline
\\
\textbf{Conversation history:} \\
\tabitem I love the taste of pasta mixed with tomato sauce and shredded cheese all over it\\
\tabitem I just love mines with chunky tomatoes mixed with meatballs and mushrooms.\\
\tabitem Olive Garden is the place that makes the best pastas. I try to visit the place as much I can with my friends\\
\tabitem  I love Olive Garden especially the original one in Orlando they opened in 1982.\\
\tabitem I think Italians have special genes that makes them so adept at preparing pastas\\
\tabitem  I believe so as well I would love to visit Argentina that is where Italian-born people reside great pasta I bet.\\
\tabitem I don't know how they do it, but no matter how hard I try to prepare one at home, mine ends up with some peculiar taste\\~\\
\textbf{\retwog{}:} I don't know how they do it but I love the taste of pasta mixed with tomato sauce and shredded cheese all over it \\\texttt{[INCONSISTENT - Dialogue repetition]}\\\hline
\textbf{\kgi{0}:} I just love the taste of pasta mixed with tomato sauce and shredded cheese all over it \\\texttt{[INCONSISTENT - Dialogue repetition]}\\\hline
\textbf{Ground truth:} The taste is so full of an abundance of taste it just can't be copied lol. \\\texttt{[INCONSISTENT - Semantically incoherent text ``taste is full of .. taste''.]}\\\hline

\end{tabular}
\caption{Some of the randomly selected instances from WoW benchmark that were evaluated by a human evaluator. The comments inside \texttt{[..]} are the feedback provided by the evaluator.}
\label{tbl.coherence_test}
\end{table*}

\subsection{Generation Quality}

Table \ref{tbl.coherence_test} shows couple of examples that were part of the set of randomly selected instances from WoW dataset and used for manual inspection. We choose these two particular instances to show when we thought the ground truth (i.e. target) is not coherent with respect to the corresponding conversation history.

In the first example, the system generated outputs were judged as coherent. We found that both \retwog{} and \kgi{0} retrieved the following passage which might have helped generation of the above output -

\begin{quote}
    \it Horseshoe Falls / Horseshoe Falls Horseshoe Falls, also known as Canadian Falls, is the largest of the three waterfalls that collectively form Niagara Falls on the Niagara River along the Canada–United States border. Approximately 90\% of the Niagara River, after diversions for hydropower generation, flows over Horseshoe Falls. The remaining 10\% flows over American Falls and Bridal Veil Falls. It is located between Terrapin Point on Goat Island in the US state of New York, and Table Rock in the Canadian province of Ontario. Section: International border.
\end{quote}

As for the ground truth, we marked it (factually) inconsistent based on the following retrieved passage -

\begin{quote}
    \it Niagara Falls / Located on the Niagara River, which drains Lake Erie into Lake Ontario, \textbf{the combined falls} have the highest flow rate of any waterfall in North America that has a vertical drop of more than . During peak daytime tourist hours, more than 168,000 m (\textbf{six million cubic feet}) of water goes over the crest of \textbf{the falls} every minute. Horseshoe Falls is the most powerful waterfall in North America, as measured by flow rate.
\end{quote}

In the second example, all three texts were marked inconsistent. Interestingly, all the items in the conversation contains subjective opinion. Consequently, all the three candidate texts also contains subjective opinion. The problem is both the systems generated texts that are almost repetition of earlier conversation. In case of the ground truth, we find that the text is semantically incoherent.

We have also submitted files that contain all instances that were used to generate the different analysis reported in Section 4.2 of the paper. These files also contains our annotations/remarks where applicable.

\end{document}